\documentclass[letterpaper, 10 pt, conference]{ieeeconf}  
\IEEEoverridecommandlockouts                              
\usepackage{amsmath}     
\usepackage{amssymb}     
\usepackage{graphicx}
\usepackage{algorithm}
\usepackage[noend]{algpseudocode}
\usepackage{dblfloatfix}
\usepackage{xcolor}
\usepackage{float}
\usepackage{booktabs}
\usepackage{multirow}
\usepackage{cite}

\usepackage[font=small]{caption}
\usepackage{subcaption}

\def\eqref#1{Eq.~(\ref{#1})}

\newcommand\etal{\emph{et al.}}

\title{\LARGE \bf Point of View: How Perspective Affects Perceived Robot Sociability}

\author{Subham Agrawal \and Aftab Akhtar \and Nils Dengler \and Maren Bennewitz
  \thanks{All authors are with the Humanoid Robots Lab, University of Bonn, Germany. M. Bennewitz is additionally with the Lamarr Institute for Machine Learning and Artificial Intelligence and the Center for Robotics, Bonn, Germany. This work has partially been funded by the German Federal Ministry of Research, Technology and Space~(BMFTR) under the Robotics Institute Germany (RIG). The paper has been accepted for publication at the 35th IEEE International Conference on Robot and Human Interactive Communication (RO-MAN 2026)
  }%
}

\begin{document}
\maketitle
\thispagestyle{empty} 
\pagestyle{empty}

\begin{abstract} 

Ensuring that robot navigation is safe and socially acceptable is crucial for comfortable human-robot interaction in shared environments. However, existing validation methods often rely on a bird's-eye (allocentric) perspective, which fails to capture the subjective first-person experience of pedestrians encountering robots in the real world. In this paper, we address the perceptual gap between allocentric validation and egocentric experience by investigating how different perspectives affect the perceived sociability and disturbance of robot trajectories. Our approach uses an immersive VR environment to evaluate identical robot trajectories across allocentric, egocentric-proximal, and egocentric-distal viewpoints in a user study. We perform this analysis for trajectories generated from two different navigation policies to understand if the observed differences are unique to a single type of trajectory or more generalizable. We further examine whether augmenting a trajectory with a head-nod gesture can bridge the perceptual gap and improve human comfort. Our experiments suggest that trajectories rated as sociable from an allocentric view may be perceived as significantly more disturbing when experienced from a first-person perspective in close proximity. Our results also demonstrate that while passing distance affects perceived disturbance, communicative social signaling, such as a head-nod, can effectively enhance the perceived sociability of the robot's behavior.

  %
  

 
\end{abstract}

\section{Introduction}
\label{sec:intro}

As robots increasingly operate in indoor and outdoor environments shared with humans, ensuring that their navigation behavior is not only safe but also socially acceptable is crucial for comfortable human-robot interaction~\cite{vega2017socially}.
To achieve this, research in the domain of social robotics often considers the humans' subjective perception of the robot's sociability~\cite{guillen2023evolution}.
Most studies evaluating perceived sociability use observations of robot behavior and interactions from a bird's-eye (allocentric) perspective~\cite{gao2022evaluation} using simplified methods, e.g., videos of the interactions~\cite{lemasurier2024comparing}.
However, this does not reflect real-world human experience.
People naturally interact from a first-person (egocentric) perspective, yet this viewpoint is rarely used to validate navigation algorithms due to practical constraints associated with live unstructured human-robot experiments~\cite{mavrogiannis2019effects,mavrogiannis2023core}.

\begin{figure}[ht]
  \centering
 \includegraphics[width=0.99\linewidth]{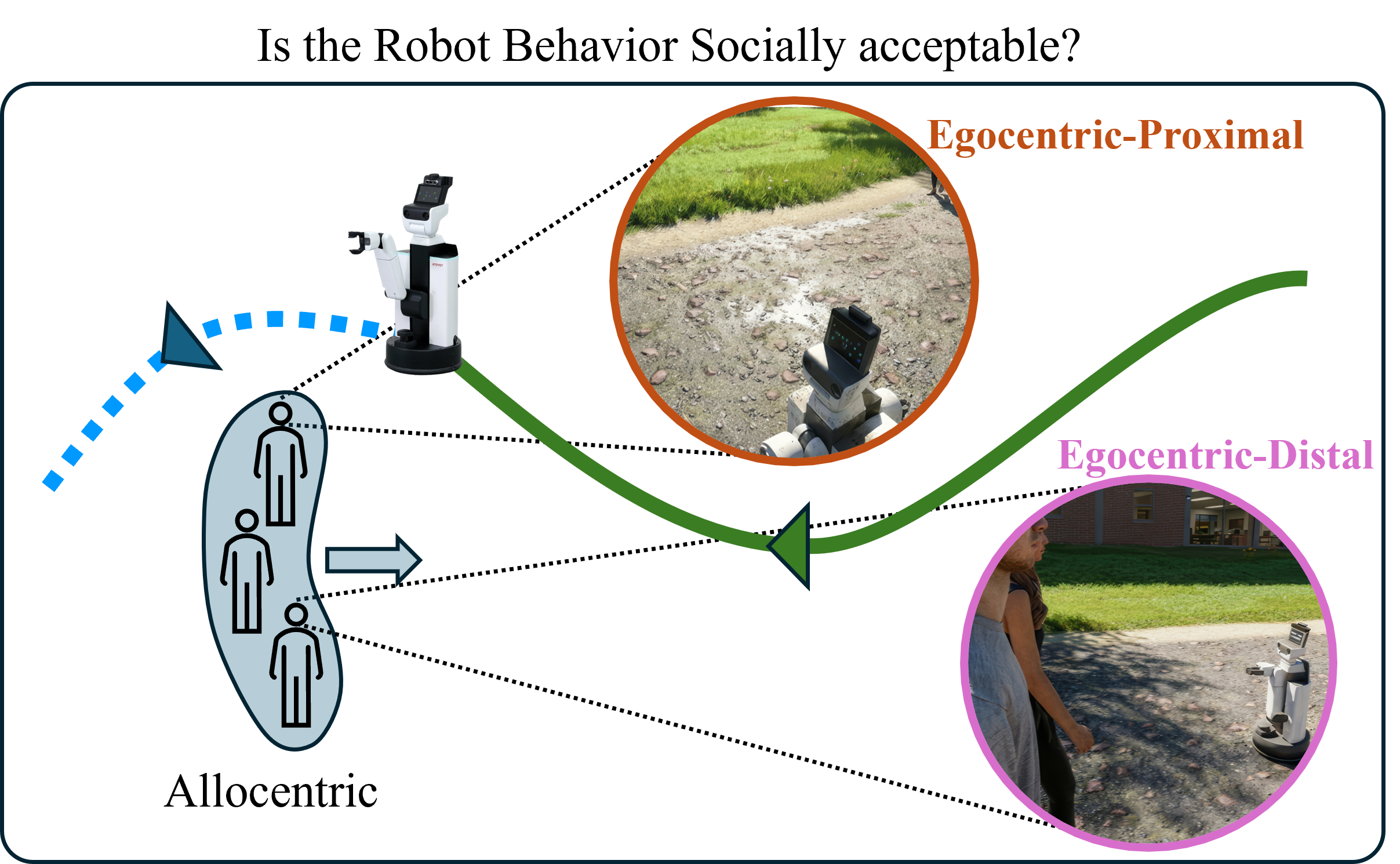}
 \caption{Example robot trajectory from different perspectives as evaluated in our user study. The \textit{allocentric} perspective is rated as socially acceptable robot behavior. The \textit{egocentric-proximal} perspective experienced by the person closest to the robot results in high perceived discomfort, whereas the \textit{egocentric-distal} perspective experienced by the person farthest away from the robot shows the robot at a comfortable distance. Based on these findings, we provide design insights for navigation strategies.}
 \label{fig:motivation}
\end{figure}

Furthermore, existing research involving human participants often relies on comparative assessments of sociability~\cite{bera2018socially,wen2022socially}. 
This approach biases the user's opinion by offering a comparative choice between two trajectories rather than explicitly quantifying whether either of them is inherently sociable.
These create a misalignment between validation and deployment: a trajectory rated as sociable from an allocentric perspective may be perceived as uncomfortable or threatening by a pedestrian experiencing it from an egocentric perspective.
Research suggests that augmenting robot behavior by integrating multi-modal signals can clarify robot intent and improve human comfort, e.g., by incorporating social gestures such as a head-nod~\cite{guillen2023evolution,heerink2006influence}.
However, there is a lack of empirical evidence both to quantify the perceptual bias on robot behavior and to determine whether a communicative augmentation is effective in closing it.

To address the perceptual gap between allocentric validation and egocentric deployment, and the lack of empirical evidence for communicative augmentation, we conduct a user study in a virtual environment.
In this study, the participants evaluate identical robot trajectories with respect to sociability and disturbance metrics from the Human-Robot Interaction Evaluation Scale (HRIES)~\cite{spatola2021perception}.
Two trajectories generated under different navigation policies are used to assess whether the observed metrics are specific to certain policy types or generalizable.
The user study questions are presented for all three perspectives: allocentric (2D overhead), egocentric-proximal (close proximity), and egocentric-distal (farther proximity), as shown in Fig.~\ref{fig:motivation}.
We further examine the effect of communicative head-nod gesture-based augmentation on perceived sociability in egocentric conditions, where bare trajectories are rated poorly.
Our results show that perceived sociability is lowest in the egocentric-proximal perspective, whereas perceived disturbance is highest from the same perspective.
Furthermore, we found that augmenting the robot's trajectory with head-nodding behavior improves participants' perceived sociability, particularly in egocentric conditions, where the robot's presence is more prominent.
Together, these findings provide guidance for designing socially aware navigation algorithms that enhance user experience and reduce perceived disturbance in close-proximity scenarios. 

To summarize, the contributions of our work are:
\begin{itemize}
    \item A user study showing that perceived sociability and disturbance differ significantly across viewpoints.
    \item An analysis demonstrating that communicative head-nod gestures improve sociability ratings across viewpoints. 
    \item Design insights for navigation strategies, such as appropriate passing distances to reduce perceived disturbance and the integration of social gestures to enhance sociability.
\end{itemize}
 

\vspace{-0.3em}
\section{Related Work}
\label{sec:related}


While social navigation demands efficient and safe robot motion in human environments, the field often overemphasizes objective metrics, i.e., navigation efficiency, success, collision avoidance, and proxemics~\cite{francis2025principles,gao2022evaluation,mavrogiannis2023core}, often overlooking the importance of subjective human feedback. The metrics offer reproducibility but fail to capture the subjective quality of a robot’s trajectory in shared space, leaving a gap between objective performance and human experience.

Metrics such as sociability, defined as how well the robot follows the social norms expected by surrounding pedestrians~\cite{luber2012socially}, are harder to operationalize.
Field studies provide opportunities to collect natural interaction data between humans and robots without any bias from instructions; however, individual encounters are not directly reproducible~\cite{gao2022evaluation}.
Additionally, large-scale field studies are resource-intensive, complex, and may expose participants to safety risks~\cite{gao2022evaluation}.
This leads to either most studies skipping subjective user ratings entirely or to testing in a scripted lab environment with a small number of participants, which limits the ecological validity of the study, as discussed by Sabanovic \etal~\cite{vsabanovic2010robots}.
Thus, there is fragmentation in social navigation evaluation methodologies without a consensus about the optimal option~\cite{francis2025principles}.

A viable middle ground to bridge the gap between fully controlled experiments and unstructured field observations is to use intercept studies, which is a methodology from survey research.
In this method, participants are stopped in public settings after a naturalistic encounter and asked questions about their experience~\cite{sudman1980improving}.
This approach has been applied in human-robot interaction to study reactions to robots in real-world deployments.
For example, Babel \etal~\cite{babel2022findings} deployed a cleaning robot at a train station and assessed pedestrians' trust and acceptance of the robot via questionnaires.
Such studies demonstrate the viability of this method for immediate, context-grounded reactions to robot behaviour.
However, such intercept-based studies do not yet evaluate trajectories specifically or compare sociability across different navigation strategies, which is the focus of the present study.
In this work, we use a virtual-reality-based environment to conduct a similar intercept-based study, thereby making it repeatable and facilitating comparison of multiple navigation strategies. 

A substantial portion of HRI user studies examines interaction from the user's perspective.
However, many of these studies are based on static robots executing simple pre-defined movements~\cite{walters2006exploratory}.
These studies effectively isolate proximity and approach direction as variables, but do not account for the dynamic, trajectory-level decisions a mobile robot makes in real time.
Thus, the results from these studies provide limited insights into how users perceive a robot executing a navigation strategy around them.
Other studies, such as those by De Heuvel \etal~\cite{de2025impact}, use both VR and 2D interfaces to examine how user preferences for robot navigation vary with immersion level, particularly highlighting the advantages of VR in improving user engagement and spatial awareness.
Similarly, \mbox{Wozniak~\etal~\cite{wozniak2023happily}} and LeMasurier \etal~\cite{lemasurier2024comparing} demonstrate that VR interfaces enable a richer, more intuitive understanding of robot motion and decision-making, particularly in complex, dynamic environments.
Our work takes inspiration to set up a similar VR interface to safely investigate perception-based effects of the perceived sociability of robot navigation strategies. 

Additionally, several studies also look at expressive social gestures and the reaction of humans to these gestures~\cite{de2023design}. Mutlu \etal~\cite{mutlu2009footing}, Baraka \etal~\cite{baraka2016enhancing}, and Scales \etal~\cite{scales2024planning} have explored LED displays, head motion, and speed modulation as communication signals.
However, the potential of these modules and gestures to modulate users' perception of robot navigation has not been investigated in depth~\cite{pascher2023communicate}.
Our work aims to look into this gap, as expressive behaviour may serve as a mechanism to compensate for navigation strategies that are technically efficient but socially ambiguous.



\section{Method}
\label{sec:method}

\subsection{Problem Statement}
Our work analyzes differences in perceived sociability and disturbance of robot navigation trajectories as experienced by pedestrians from different perspectives. Specifically, we consider three different perspectives: \textit{Allocentric (2D)}, \textit{Egocentric-Proximal~(3D Near)}, and \textit{Egocentric-Distal~(3D Far)}. The primary objective of our research is to address the following questions:

\begin{itemize}
    \item \textbf{RQ1:} Does the perceived sociability of a robot trajectory differ across different perspectives: allocentric~(2D), egocentric-proximal, and egocentric-distal?
    \item \textbf{RQ2:} Does modifying robot behavior by adding a social gesture improve the sociability of the trajectory?
    \item \textbf{RQ3:} Can social navigation-based sociability thresholds derived using evaluations in simulation be valid for egocentric settings?
\end{itemize}


\begin{figure*}[ht]
    \centering
    \begin{subfigure}{0.32\textwidth}
        \centering
        \includegraphics[width=\linewidth]{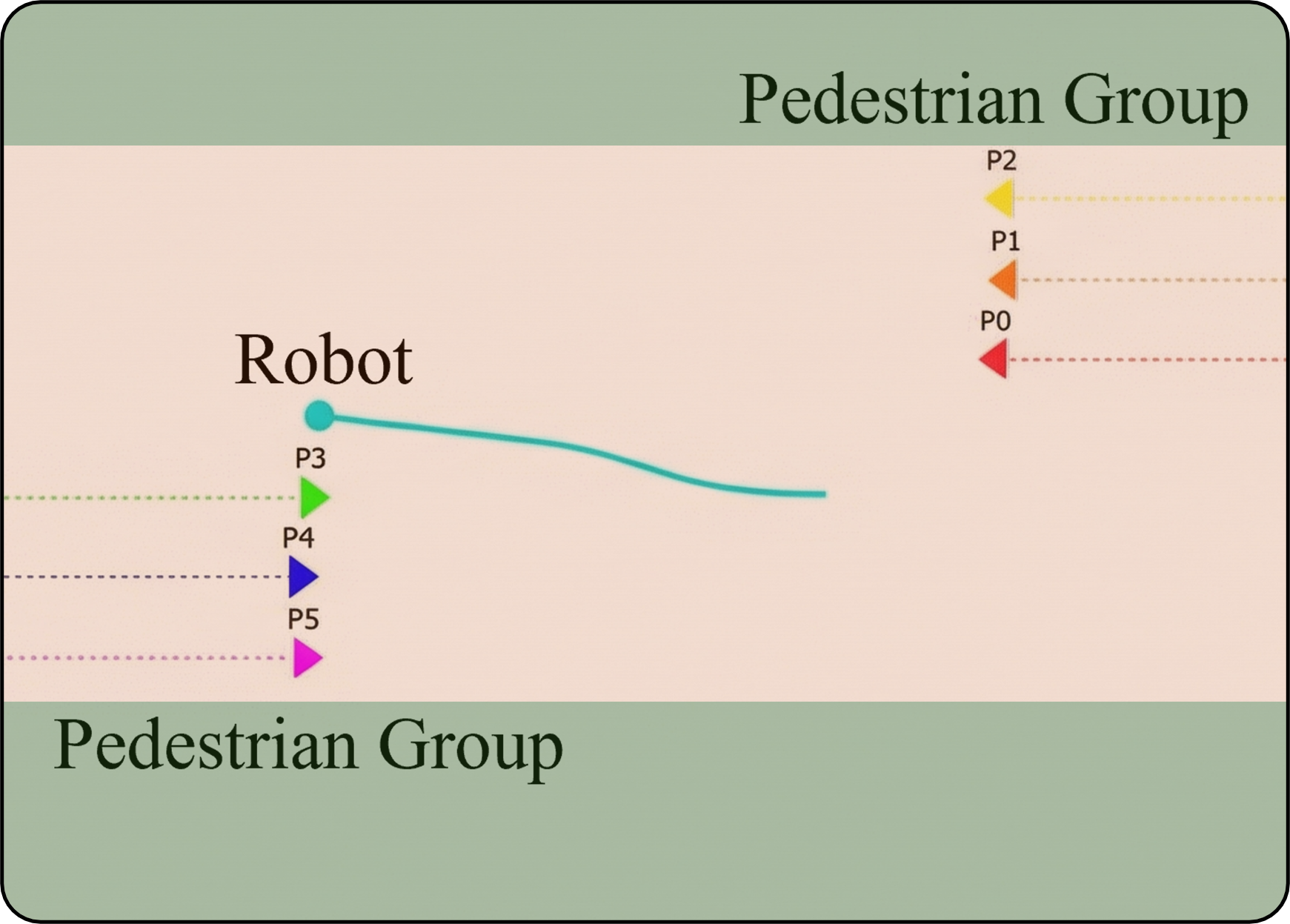}
        \caption{Allocentric}
    \end{subfigure}
    \hfill
    \begin{subfigure}{0.32\textwidth}
        \centering
        \includegraphics[width=\linewidth]{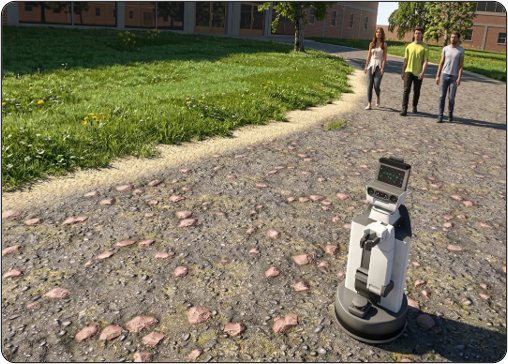}
        \caption{Egocentric-Proximal}
    \end{subfigure}
    \hfill
    \begin{subfigure}{0.32\textwidth}
        \centering
        \includegraphics[width=\linewidth]{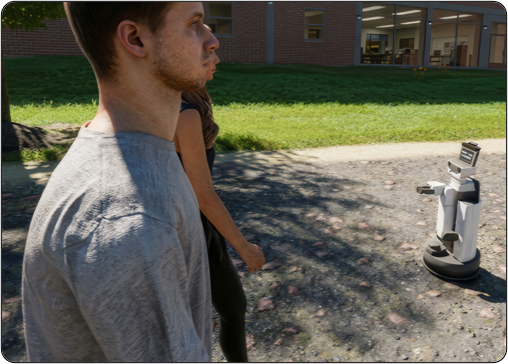}
        \caption{Egocentric-Distal}
    \end{subfigure}
    \caption{The different perspectives of the same trajectory presented to participants: (a) allocentric perspective showing the bird's-eye view of the trajectories representative of simulation environments, (b) egocentric-proximal perspective of the pedestrian P3 from (a), and (c) egocentric-distal perspective of the pedestrian P5 from (a).}
    \label{fig:perspectives}
\end{figure*}

\subsection{Virtual Reality Interface}
\label{sec:vr_interface}
To ensure experimental control and reproducibility, we develop an immersive virtual reality (VR) interface using the Unity game engine for the Meta Quest~3. 
This setup enables controlled, repeatable human-robot experiments without robot platform errors, safety concerns, or environmental disturbances inherent to real-world deployments. 
Although conducting the study with a real robot might amplify pedestrians’ responses due to the robot’s consequential sound, physical presence, and perceived safety risks, real-world setups are also less repeatable and these cues could also make it harder to determine whether the observed effects are driven by perspective itself.
Using VR allows us to reduce these additional influences and isolate the effect of perspective on subjective sociability and disturbance ratings.
The VR scene depicts a shared walking space in which two pedestrian groups approach one another from opposite lanes. 
Each group consists of three individuals arranged in a V-shaped formation, reflecting the most commonly observed configuration in naturally occurring pedestrian groups~\cite{moussaid2010walking}. 
The groups walk toward each other at a speed of $0.85~m/s$ to reduce motion sickness in VR~\cite{page2023identifying}, consistent with the lower limit of the range of average pedestrian walking speeds reported in literature~\cite{knoblauch1996field}.

\subsection{Perspectives}
\label{sec:perspectives}
A key contribution of this study is the systematic comparison of how the same robot navigation behavior is perceived from different viewpoints. We consider three perspective conditions, as shown in Fig.~\ref{fig:perspectives}:

\begin{itemize}
    \item \textit{Allocentric (2D)}: The participant observes a top-down view of the entire interaction. This perspective is representative of simulation-based navigation evaluations and reflects the viewpoint most commonly used for designing and tuning navigation policies. It also provides a baseline for sociability and disturbance ratings in the current evaluation methodology.

    \item \textit{Egocentric-Proximal}: The participant views the interaction from the perspective of the pedestrian walking closest to the robot's movement trajectory. This condition captures the most immediate and spatially closest experience of the robot's trajectory.

    \item \textit{Egocentric-Distal}: The participant views the interaction from the perspective of the pedestrian positioned furthest from the robot within the group. This condition reflects a more peripheral, but still embodied, experience of the navigation event. Within the egocentric subcondition, this perspective allows isolation of passing distance as a variable and helps answer the question of how far a robot should pass for minimal disturbance.
\end{itemize}

\subsection{Robot and Navigation Trajectories}
\label{sec:trajectories}

\begin{figure}[ht]
  \centering
 \includegraphics[width=0.99\linewidth]{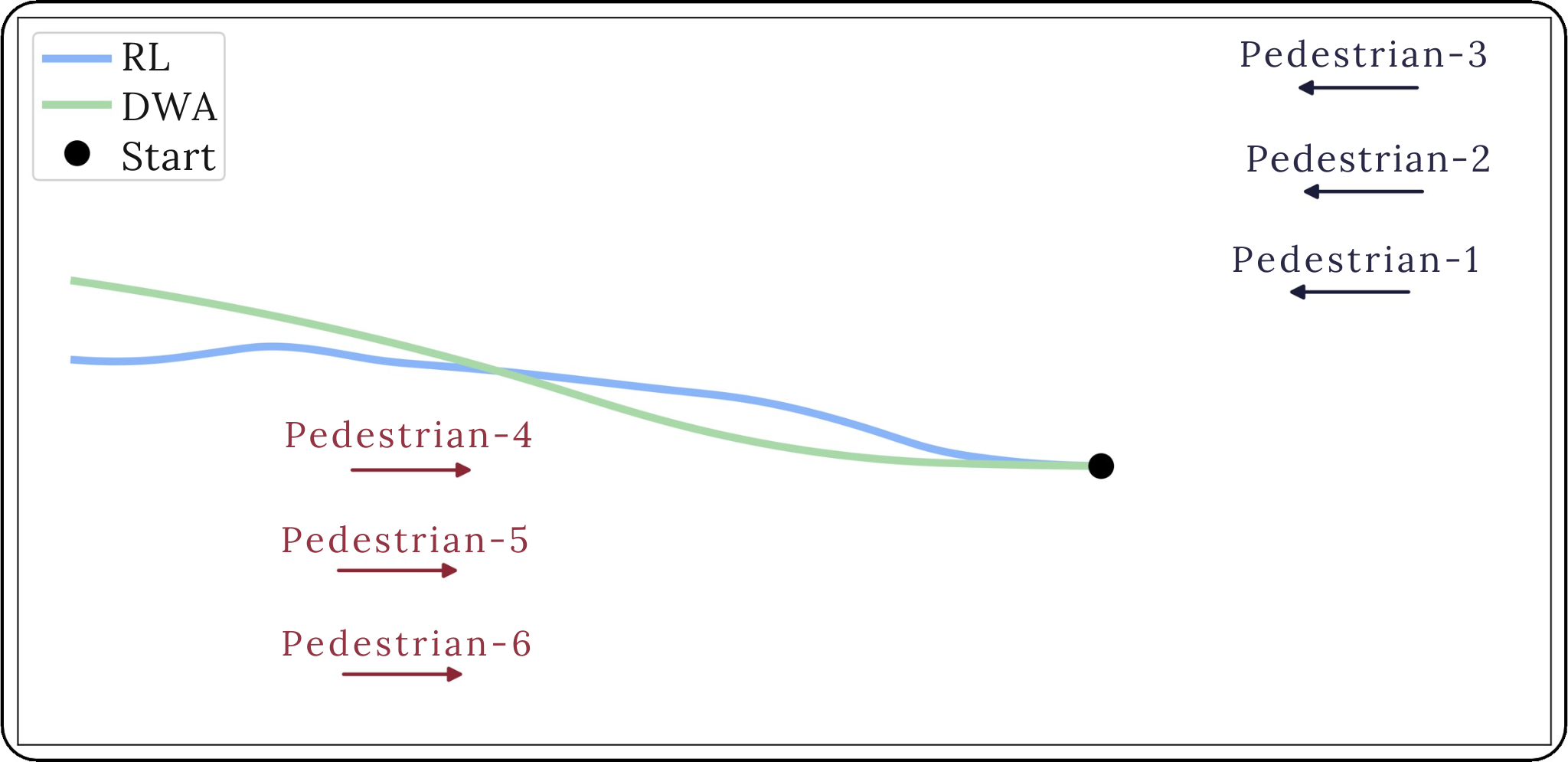}
 \caption{DWA and RL-based trajectories used for the user study. The DWA trajectory (green) is smooth, whereas the RL-based trajectory~(blue) changes lane early. Both trajectories have similar passing distances ($ 0.95\,m $) to the closest pedestrian. The passing distance from the farthest pedestrian ($ 2.53\,m $) is also similar.}
 \label{fig:trajectories}
\end{figure}

We use the Toyota Human Support Robot (HSR)~\cite{yamamoto2019development} to navigate through the group of pedestrians.
It is a mobile robot designed to assist people in everyday environments and has been used in multiple studies on human–robot interaction and social navigation~\cite{agrawal2024evaluating,agrawal2025peroi,kim2018social}.
The HSR follows predefined trajectories generated by two distinct navigation policies: the classical Dynamic Window Approach~(DWA)~\cite{fox2002dynamic}, and a Reinforcement Learning (RL) based social navigation policy~\cite{agrawal2024evaluating}. 
These policies produce two distinct robot trajectories, as shown in Fig.~\ref{fig:trajectories}, enabling a direct comparison of their perceived sociability across perspectives. 
Using two navigation policies was essential to determine whether the observed differences in sociability and disturbance are specific to one trajectory type or also apply to other types, thereby making these observations more generalizable.
Additionally, we evaluate in the study an augmented RL policy-based trajectory with a comforting social gesture to investigate whether this improves the perceived sociability.
To keep the overall time of the user study within a reasonable limit for participants, only the RL-based trajectory was augmented. In the following, we introduce the three navigation policies in more detail.

\subsubsection{DWA}
We use a standard DWA local planner, fine-tuned with a 9-second future-planning window. 
This timeframe allows the robot to account for pedestrian motion further in advance, resulting in more proactive path planning. 
Fig.~\ref{fig:trajectories} shows the DWA-based trajectory demonstrating a smooth trajectory behavior due to the selected timeframe.
This policy serves as a well-established baseline representative of conventional navigation approaches.

\subsubsection{RL based Policy}
The RL-based social navigation policy generates trajectories learned by the robot in a simulation environment~\cite{agrawal2024evaluating}. 
This policy serves as a representative of machine-learning-based policies in the literature. 
Fig.~\ref{fig:trajectories} shows that the RL-based trajectory adjusts its lane at an earlier point than the DWA-based trajectory.
Both trajectories have a minimum passing distance of $ 0.95\,m $ to the closest pedestrian.

\subsubsection{RL-based Policy (Augmented)}
This condition uses an identical trajectory to the RL-based policy but augments the robot's motion with a communicative head-nod social gesture. 
This gesture is directed specifically at the participant from whose perspective the scene is viewed. 
Such behavior ensures that, under both egocentric conditions, the robot maintains an explicit social acknowledgment of the user. 
By keeping the underlying path constant while varying only the expressive behavior, we isolate the specific effect of social signaling on the perceived sociability of the trajectory across different perspectives.
The gesture is triggered as the robot approaches and is about $ 2.5\,m$  away from the participant's position. 
The head-nod consists of a pitch rotation of~$45^\circ$. 
To ensure the gesture is perceptible and socially appropriate, the robot reduces its velocity to~60\% of its normal navigation speed during execution. 
Once the head returns to its neutral orientation, the robot resumes its predefined velocity and continues along its path.

\subsection{Experimental Design}
\label{sec:design}
We employ a within-subjects design to evaluate the impact of navigation policies and viewing perspectives on perceived social acceptability. 
The experiment follows a $3 \times 3$ factorial design with two independent variables:
\begin{itemize}
    \item \textit{Navigation Condition}: DWA, RL-based Policy, and RL-based Policy (Augmented).
    \item \textit{Viewpoint}: Allocentric, Egocentric-Proximal, and Egocentric-Distal.
\end{itemize}

This produces 9 unique experimental conditions per participant. 
An a priori power analysis conducted in G*Power for a repeated-measures ANOVA with medium effect size \mbox{$f = 0.22$}, $\alpha = 0.05$, and power~$0.80$ across 9 measurements indicates a minimum required sample size of 19 participants. 
To mitigate carryover effects and ensure balanced condition orderings, trial sequences are randomized per participant using a Williams Latin Square design~\cite{williams1949experimental}.

\begin{figure*}[ht]
    \centering
    \begin{subfigure}{0.48\textwidth}
        \centering
        \includegraphics[width=\linewidth]{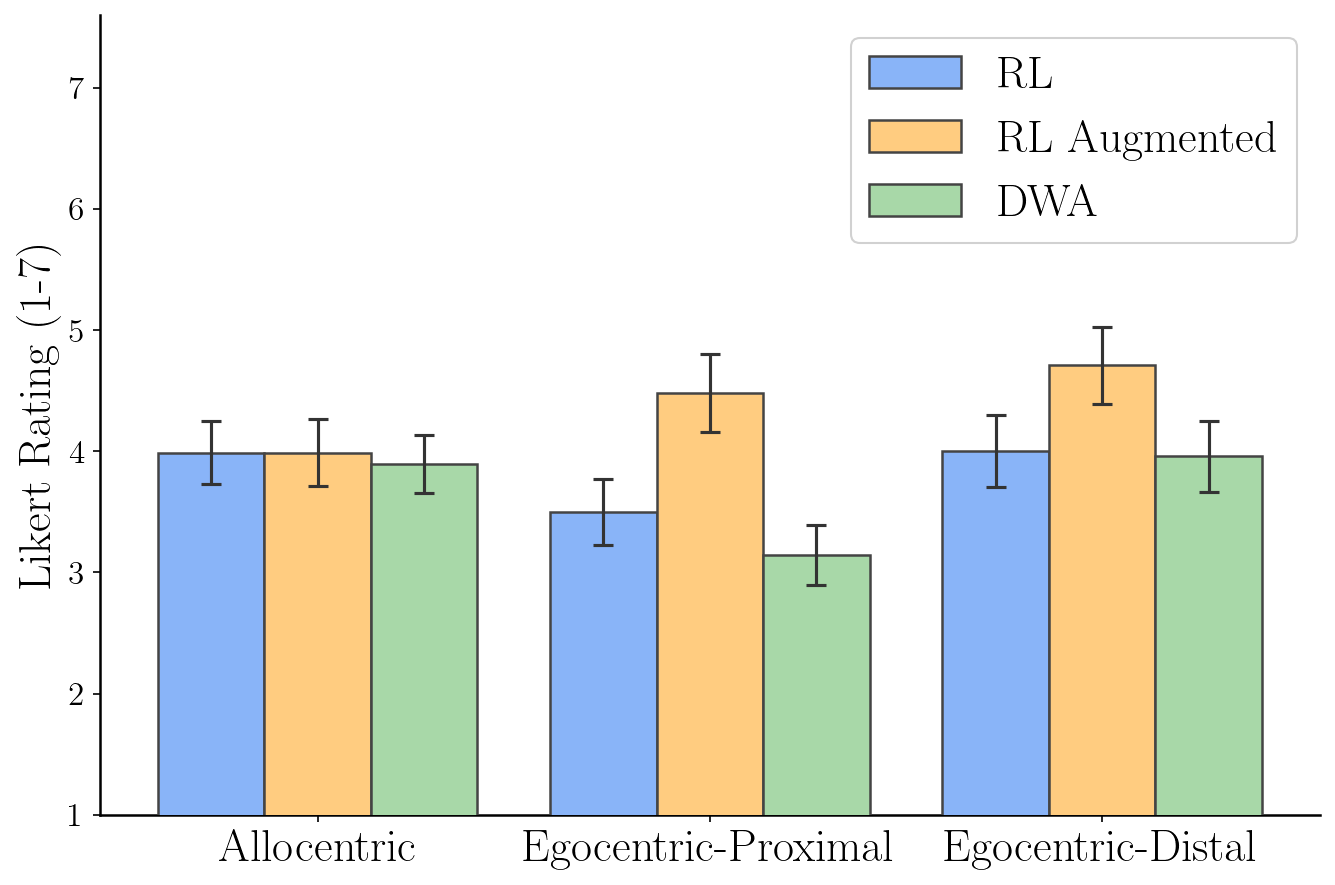}
        \caption{Sociability Ratings}
    \end{subfigure}
    \hfill
    \begin{subfigure}{0.48\textwidth}
        \centering
        \includegraphics[width=\linewidth]{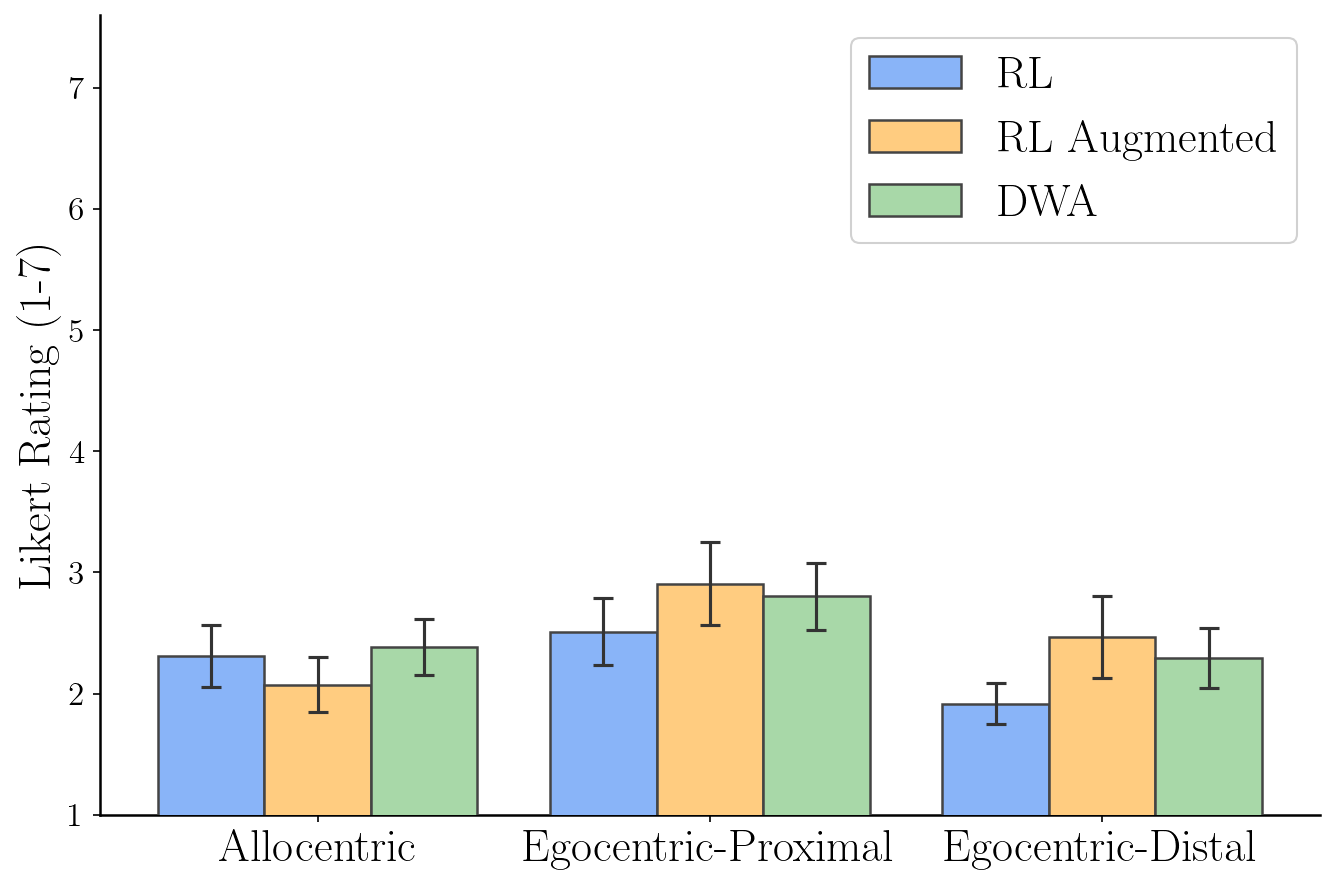}
        \caption{Disturbance Ratings}
    \end{subfigure}
    \caption{The sociability and disturbance rating results from 24 participants. a) Sociability ratings show the highest sociability for the egocentric-distal case, but the differences amongst viewpoints are not significant. In all cases, the RL-based trajectory augmented with head-nodding received the highest sociability score. b) Disturbance ratings show the highest disturbance for the egocentric-proximal case, which has a significant difference compared to the other two viewpoints. The augmentation did not cause any significant effects on these ratings across viewpoints.}
    \label{fig:results}
\end{figure*}

\begin{table*}[ht]
\centering

\begin{tabular}{@{}llccc@{}}
\toprule
\textbf{Construct} & \textbf{Comparison} & \textbf{Test Statistic} & \textbf{\textit{p}-value} & \textbf{Effect Size ($g$)} \\ \midrule

\multicolumn{5}{@{}l}{\textbf{1. Omnibus Tests (Overall Main Effects)}} \\ \midrule
Sociability & Main Effect of Condition (9 levels) & $F(8, 184) = 5.57$ & \textbf{0.0007} & $\eta_p^2 = 0.094$ \\
Disturbance & Main Effect of Condition (9 levels) & $\chi^2(8) = 20.49$ & \textbf{0.0086} & -- \\ \midrule

\multicolumn{5}{@{}l}{\textbf{2. Main Effects of Viewpoint (FDR Corrected)}} \\ \midrule
\multirow{3}{*}{Sociability} 
 & Egocentric-Distal vs. Egocentric-Proximal & $t = 4.40$ & \textbf{0.0006} & $0.44$ \\
 & Allocentric vs. Egocentric-Proximal & $t = 1.50$ & 0.2136 & $0.23$ \\
 & Allocentric vs. Egocentric-Distal & $t = -1.28$ & 0.2136 & $-0.22$ \\ \addlinespace
\multirow{3}{*}{Disturbance} 
 & Egocentric-Distal vs. Egocentric-Proximal & $W = 34.0$ & \textbf{0.0084} & $-0.48$ \\
 & Allocentric vs. Egocentric-Proximal & $W = 56.5$ & \textbf{0.0359} & $-0.45$ \\
 & Allocentric vs. Egocentric-Distal & $W = 98.0$ & 0.8082 & $0.03$ \\ \midrule

\multicolumn{5}{@{}l}{\textbf{3. Targeted Simple Effects (Uncorrected)}} \\ \midrule
\multirow{3}{*}{Sociability} 
 & RL + Head Nod vs. RL (Proximal) & $t = 2.44$ & \textbf{0.0228} & $0.66$ \\
 & RL + Head Nod vs. RL (Distal) & $t = 2.10$ & \textbf{0.0465} & $0.46$ \\
 & DWA vs. RL (Proximal) & $t = -2.86$ & \textbf{0.0089} & $-0.27$ \\ \addlinespace
\multirow{3}{*}{Disturbance} 
 & RL + Head Nod vs. RL (Proximal) & $W = 91.5$ & 0.4134 & $0.26$ \\
 & RL + Head Nod vs. RL (Distal) & $W = 63.5$ & 0.1238 & $0.42$ \\
 & DWA vs. RL (Proximal) & $W = 52.0$ & 0.2548 & $0.21$ \\ \bottomrule
\end{tabular}
\caption{Statistical Analyses for Sociability and Disturbance. Significant $p$-values ($< 0.05$) are presented in bold. Sociability omnibus test uses Greenhouse-Geisser corrected $p$-value. Results show that viewpoint significantly affects both perceived sociability and disturbance, whereas behavioral enhancements, such as head nodding, increase sociability without significantly affecting disturbance levels.}
\label{tab:stats_summary}
\end{table*}

\begin{table}[ht]
\centering
\resizebox{\columnwidth}{!}{
\begin{tabular}{llcccc}
\toprule
& & \multicolumn{2}{c}{\textbf{Sociability}} & \multicolumn{2}{c}{\textbf{Disturbance}} \\
\cmidrule(lr){3-4} \cmidrule(lr){5-6}
\textbf{Perspective} & \textbf{Trajectory} & \textbf{Mean} & \textbf{SD} & \textbf{Mean} & \textbf{SD} \\ 
\midrule
\multirow{3}{*}{Allocentric} 
 & DWA & 3.90 & 1.17 & 2.39 & 1.14 \\
 & RL-based & 3.99 & 1.26 & 2.31 & 1.25 \\
 & RL-based + Head Nod & 3.99 & 1.36 & 2.07 & 1.11 \\ 
\addlinespace
\multirow{3}{*}{Egocentric-Distal} 
 & DWA & 3.96 & 1.42 & 2.29 & 1.21 \\
 & RL-based & 4.00 & 1.45 & 1.92 & 0.82 \\
 & RL-based + Head Nod & 4.71 & 1.57 & 2.47 & 1.65 \\ 
\addlinespace
\multirow{3}{*}{Egocentric-Proximal} 
 & DWA & 3.15 & 1.22 & 2.80 & 1.36 \\
 & RL-based & 3.50 & 1.33 & 2.51 & 1.35 \\
 & RL-based + Head Nod & 4.48 & 1.59 & 2.91 & 1.68 \\ 
\bottomrule
\end{tabular}
}
\caption{Sociability and Disturbance across Perspective and Trajectory conditions. Means and Standard Deviations are reported. Results indicate that the RL-based trajectory with a head nod achieves the highest sociability ratings across all perspectives. The egocentric-proximal condition is the least sociable and most disturbing perspective, regardless of the trajectory used.}
\label{tab:result_summary}
\end{table}

\subsection{User Study}
\label{sec:userstudy}
At the start of the study, participants receive an information sheet and provide written informed consent. 
Demographic data are collected alongside a baseline assessment of attitudes toward robots using the Negative Attitudes toward Robots Scale (NARS)~\cite{nomura2006measurement}. 
The experimental procedure then proceeds as follows:

\begin{itemize}
    \item \textit{Familiarization}: Participants complete one introductory trial to get familiar with the Meta Quest~3 and the virtual environment. They also complete questionnaires to gain experience with the interface used to collect responses during the trials.
    \item \textit{Experimental Trials}: Participants observe the robot navigating across all 9 counterbalanced conditions. After each trial, a 7-point Likert scale questionnaire (``Totally Agree'' to ``Not at All'') appears within the VR interface, capturing subjective ratings of sociability and disturbance across 8 questions. To evaluate the overall social comfort of the robot's navigation behavior, we selected two distinct subscales from the HRIES questionnaire to be presented as questions to the participants - \textbf{Sociability} (positive valence: warm, trustworthy, likeable, and friendly) and \textbf{Disturbance} (negative valence: scary, creepy, uncanny, and weird). This dual-scale approach mitigates acquiescence bias and allows independent evaluation of whether specific navigation trajectories and views enhance social presence, mitigate disturbing responses, or both.  
    \item \textit{Post-Study Evaluation}: After the completion of the trials, participants complete a final questionnaire on behavioral preferences and provide qualitative feedback.
\end{itemize}

We recruited a total of 24 participants (\mbox{$M_{\text{age}} = 27.3$ years}, $SD = 3.6$) for the user study.
Participants reported moderate prior experience with mobile robots and exhibited a neutral baseline attitude toward robots as measured by the NARS. 
All participants reported normal or corrected-to-normal vision.

\vspace{-0.3em}
\section{Results and Discussion}
\label{sec:exp}

\subsection{Scale Reliability and Statistical Tests}

Before conducting the primary analysis of the results, the internal consistency of the sociability and disturbance scales is assessed. 
Both scales demonstrate excellent consistency and reliability, with Cronbach's alpha for sociability (\mbox{$\alpha = 0.915$}) and disturbance (\mbox{$\alpha = 0.91$}) above the acceptable threshold of 0.7.

To determine the appropriate statistical tests, the data was analyzed for normality using Shapiro-Wilk tests~\cite{shapiro1965analysis} and for sphericity using Mauchly's test~\cite{mauchly1940significance}. 
Preliminary results revealed that sociability data was normally distributed, thus the parametric One-Way Repeated Measures (RM) ANOVA was used for statistical significance. 
Additionally, sociability data significantly violated the sphericity assumption, thus the Greenhouse-Geisser correction was applied to the degrees of freedom. 
The disturbance data did not meet the normality and sphericity thresholds in the results, and thus, the nonparametric Friedman test was used to analyze this category.

\subsection{Main Effects and Post-Hoc Comparisons}

The Greenhouse-Geisser corrected RM ANOVA results showed a statistically significant main effect of condition on perceived sociability (\textbf{$\eta_p^2 = 0.094, p = 0.0007$}). 
The Friedman test results indicated a statistically significant difference in perceived disturbance across the 9 experimental conditions of navigation and viewpoints (\textbf{$\chi^2 = 20.49, p = 0.0086$}).
Together, these omnibus results indicate that the combination of the robot's navigational trajectory and the user's viewing perspective significantly affects both perceived sociability and disturbance. 
Given these significant overall effects, targeted post-hoc comparisons were conducted to isolate the specific impacts of viewpoint and trajectory design.

To evaluate the main effects of viewing perspective, pairwise comparisons were conducted across the three viewpoints.
The sociability and disturbance ratings are shown in Tab.~\ref{tab:result_summary} and Fig.~\ref{fig:results}, and the associated statistical analyses are shown in Tab.~\ref{tab:stats_summary}.
The p-values of these tests were adjusted using the False Discovery Rate (FDR) correction to account for multiple comparisons. 
For sociability, paired t-tests revealed that the \textbf{egocentric-distal viewpoint was rated as significantly more sociable than the egocentric-proximal viewpoint ($p < 0.01$)}. 
However, there were no significant differences in sociability ratings between the allocentric viewpoint and either of the egocentric viewpoints.
For disturbance, Wilcoxon signed-rank tests showed that the \textbf{egocentric-proximal viewpoint was significantly more disturbing than both the allocentric ($p < 0.05$) and egocentric-distal ($p < 0.01$) viewpoints}.
These results show that users generally experience a significant decrease in social comfort when a robot operates in close physical proximity. While positive sociability feelings do not show a significant difference, negative disturbance feelings are perceived more strongly by users in closer proximity. 

\subsection{Impact of Head-Nod and Trajectory Design}

Because the egocentric-proximal perspective exhibited the worst sociability and the highest disturbance, targeted simple-effects testing was conducted to determine whether specific strategies could mitigate this issue.
To this end, we evaluated the ratings for the RL-based policy and the augmented RL-based policy. 
In the egocentric-proximal condition, \textbf{augmenting the trajectory with head-nod significantly increased sociability} compared to the standard RL-based policy \textbf{($p < 0.05$)}. 
In the egocentric-distal condition, although sociability improved, the difference was not statistically significant. 
Interestingly, head-nod-based augmentation of the trajectory did not significantly alter perceived disturbance in either egocentric-proximal or egocentric-distal conditions.
This shows that while augmenting the trajectory with a socially communicative gesture could improve the perceived sociability of the robot's behavior, it does little to mitigate the uneasiness associated with the passing distance.
Furthermore, different navigation algorithms without head-nod augmentation were also compared to understand the effect of the timing of lane change on sociability and disturbance 
For the egocentric-proximal condition, \textbf{the RL-based trajectory was rated as significantly more sociable than the traditional DWA trajectory ($p < 0.05$)}. 
However, there was no significant difference in perceived disturbance between the two trajectories. 
This suggests that the disturbance rating is more likely to be influenced by the distance at which the robot passes the user, whereas the perceived sociability is influenced by the timing of the lane change in this case.
Further studies are required to isolate and verify this finding across different scenarios. 

\subsection{NARS Subscale Effects}

The NARS responses from participants were collected to assess whether they held preconceived opinions about robots and to analyze the effect these might have on their ratings. 
The regression analyses, as shown in Tab.~\ref{tab:NARS}, revealed that, among the three subscales of the NARS questionnaire, only the Negative Attitudes toward Situations of Interaction (S1) subscale significantly predicted sociability in human-robot interaction.

\begin{table}[ht]
\centering
\resizebox{\columnwidth}{!}{
\begin{tabular}{llccccc}
\toprule
\textbf{Predictor} & \textbf{Outcome} & \textbf{b (Slope)} & \textbf{r} & \textbf{R²} & \textbf{p-value} \\
\midrule
\multirow{2}{*}{S1} 
 & \textbf{Sociability} & \textbf{-0.1527} &\textbf{ -0.6226 }& \textbf{0.3877} & \textbf{0.0012} \\
 & Disturbance & 0.0174 & 0.0876 & 0.0077 & 0.6840 \\ 
\addlinespace
\multirow{2}{*}{S2} 
 & Sociability & -0.0829 & -0.3208 & 0.1029 & 0.1264 \\
 & Disturbance & 0.0082 & 0.0393 & 0.0015 & 0.8553 \\ 
\addlinespace
\multirow{2}{*}{S3} 
 & Sociability & -0.1573 & -0.3135 & 0.0983 & 0.1357 \\
 & Disturbance & 0.0546 & 0.1344 & 0.0181 & 0.5314 \\
\bottomrule
\end{tabular}
}
\caption{Regression Analysis for Sociability and Disturbance across NARS Subscales. Only the S1 subscale explains Sociability scores, which means if participants had situational anxiety about robots, then they rated the interaction with robot as less sociable.}
\label{tab:NARS}
\end{table}

Results explain $39\%$ of the variance ($\mbox{R² = 0.388}, \mbox{p = 0.001}$), indicating that \textbf{participants who expressed greater anxiety about robot interactions rated the robot as significantly less sociable}.
In contrast, no significant relationships were found between the subscales and disturbance, suggesting that perceived disturbance was not influenced by pre-existing attitudes toward robots.
The subscale S2 (Negative Attitudes toward Social Influence of Robots) and S3 (Negative Attitudes toward Emotions in Interaction) were not significant predictors for either sociability or disturbance. 
The results indicate that perceived sociability is primarily shaped by situational anxiety (S1), whereas disturbance is driven by the robot's behavior rather than individual attitudes.

\subsection{Summary}
In summary, the results from the user study answer our research questions and indicate that perceived sociability is not significantly different between the allocentric and egocentric-proximal perspectives.
However, the perceived disturbance differs significantly, with the allocentric perspective yielding a lower disturbance rating than the egocentric-proximal perspective.
Our user study suggests that sociability metrics from allocentric perspective-based simulations are not valid in egocentric settings, as they lead to greater perceived disturbance, especially in close-proximity situations.
Finally, modifying robot behavior by augmenting the trajectory with a head-nod gesture improves the perceived sociability.
Thus, future work should adopt a dual strategy that uses social gestures to make the robot's behavior appear more social while maintaining as large a passing distance as possible to minimize inherent discomfort from the robot's physical presence.

\vspace{-0.3em}
\section{Conclusion}
\label{sec:conclusion}

In this paper, we present a novel VR-based user study to explore the impact of different perspectives: allocentric~(2D overhead), egocentric-proximal (close proximity), and egocentric-distal
(farther proximity), on the perceived sociability and disturbance caused by robot navigation behavior.
By comparing across all three perspectives, we identified where allocentric evaluations differ from paths considered sociable from a first-person perspective, and whether distance influences perceived sociability from an egocentric viewpoint.

The results of our user study indicate that perceived sociability is lowest in the egocentric-proximal perspective, while perceived disturbance is highest for the same viewpoint.
This shows that trajectories that appear sociable under the allocentric conditions used to design and validate navigation algorithms may be perceived as significantly more disturbing when experienced from a first-person perspective.
Additionally, using the head-nod gesture significantly improved sociability ratings, highlighting the importance of social signaling in close-proximity navigation.
Thus, while the passing distance primarily drives disturbance, sociability can be actively improved through communicative behavior.
Comparisons between different robot trajectories also indicate that perceived sociability may be higher for navigation behavior that leads to an earlier lane change, whereas perceived disturbance may be affected by the distance between the robot and the pedestrian. These findings provide valuable insights for improving the sociability of human aware navigation strategies.

\bibliographystyle{IEEEtran}

\bibliography{bibliography}

\end{document}